# Unsupervised domain adaptation via double classifiers based on high confidence pseudo label

*Abstract*—Unsupervised domain adaptation (UDA) aims to solve the problem of knowledge transfer from labeled source domain to unlabeled target domain. Recently, many domain adaptation (DA) methods use centroid to align the local distribution of different domains, that is, to align different classes. This improves the effect of domain adaptation, but domain differences exist not only between classes, but also between samples. This work rethinks what is the alignment between different domains, and studies how to achieve the real alignment between different domains. Previous DA methods only considered one distribution feature of aligned samples, such as full distribution or local distribution. In addition to aligning the global distribution, the real domain adaptation should also align the meso distribution and the micro distribution. Therefore, this study propose a double classifier method based on high confidence label (DCP). By aligning the centroid and the distribution between centroid and sample of different classifiers, the meso and micro distribution alignment of different domains is realized. In addition, in order to reduce the chain error caused by error marking, This study propose a high confidence marking method to reduce the marking error. To verify its versatility, this study evaluates DCP on digital recognition and target recognition data sets. The results show that our method achieves state-of-the-art results on most of the current domain adaptation benchmark datasets.

## I. Introduction

IN recent years, deep learning methods have achieved impressive success in many fields. For example: computer version[6], speech recognition[8], sentiment analysis[10], fault diagnosis[11-13] etc. however, all these methods have a premise, that is to train a deep network model, and need a large number of labeled samples. This cost is high and almost impossible in many fields. For the samples with different distributions, the performance of the model trained in the source domain are directly applied to the target domain decreases significantly. Domain adaptation (DA) aims to solve this problem. At present, unsupervised domain adaptation (UDA) is the most popular and practical research. Unsupervised domain adaptation is transferring the knowledge learned from labeled source domain samples to unlabeled target domain to migrate the discrepancy between source domain and target domain [4].

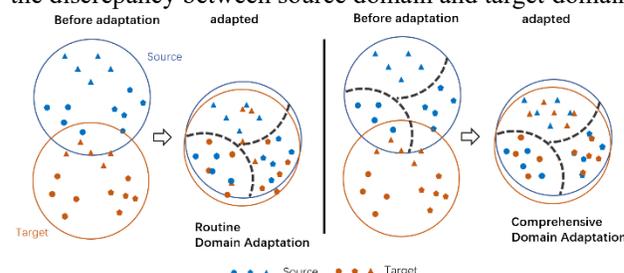

Fig. 1. Left routine subdomain adaptation might lose some meso and micro distribution information. Right comprehensive subdomain adaptation can exploit structure information for all classes and samples.

The previous unsupervised domain adaptation is divided into instances-based, feature-based, adversarial-based and model-based. Instances-based method refers to use a specific weight adjustment strategy, and different weights are applied to different samples to realize reduce the weighted training error on different distribution domains[16]. The feature-based method eliminates the difference between different domains by feature transformation[18]. Adversarial-based method refers to introduce adversarial technology inspired by generative adversarial nets (GAN) [19] to find transferable representations[20]. According to the current research results, the effect based on deep neural network is the best, which is mainly due to deep neural network has great advantages in feature extraction. The above methods mainly achieve domain alignment by reducing the global distribution difference. However, in the process of global distribution alignment, it does not mean that the classes have been aligned. Recent studies not only focus on the overall distribution discrepancy, but also focus on the local distribution discrepancy. Due to global distribution alignment, the features between classes cannot be extracted. The features between classes play a key role in the recognition and classification of samples. Currently, the most popular local alignment methods label the samples first, and then achieve class alignment by pseudo label samples. For example, [21] label the samples in the target domain by clustering algorithm, which combines the sample distribution structure of the target domain, [22] improves the accuracy of false labels by applying a certain threshold value, [23] enhances the robustness of the network by using multiple classifiers and

This work was supported by the National Natural Science Foundation of China (51475334), the National Key Research and Development Program of China (No. 2018YFE0105000, No. 2018YFB1305304) and the Shanghai Municipal Commission of Science and Technology (19511132100). (Corresponding author: Kuo-Yi Lin.) Huihuang Chen, Li Li, Jie Chen and Kuo-Yi Lin are with the College of Electronic and Information Engineering, Tongji University, Shanghai 201804, China e-mail: (see 19603@tongji.edu.cn; chenhuihuang@tongji.edu.cn; lili@tongji.edu.cn; chenjie206@tongji.edu.cn). Kuo-Yi Lin and Li Li are with the Institute of Intelligent Science and Technology, Tongji University, Shanghai 201203, China e-mail: (see 19603@tongji.edu.cn; lili@tongji.edu.cn).

random initial parameters. [25]label the samples directly through the source domain training model, without accessing source data and labels for target data, which helps protect the privacy. There has been some improvement in public data sets. These methods have made some progress in the benchmark dataset.

But we need to rethink domain adaptation. What is true domain alignment? As shown in Figure 1, the traditional local alignment is shown on the left side of Figure 1. We can see that in the process of domain alignment, the distribution between classes (called meso distribution) and the distribution between different samples of the same class (micro distribution) are not considered. How to understand meso distribution and micro distribution. We can think that the macro distribution is the overall distribution of two domains, the meso distribution is the distribution state between different classes, and the micro distribution is the distribution state of different samples between the same class. Combined with the actual situation, the overall distribution of cups and bottles in the sample space is closer than that of cups and computers, and for the same class, the samples with the same shooting angle slightly adjusted in other aspects are closer than those with huge differences in shape and color in the sample space. In the process of marking, the wrong mark will cause chain error, which reduces the effect of alignment. Based on these considerations, we think that if we want to achieve all-round domain adaptation of source domain and target domain, this study should align not only the global distribution, but also the meso distribution and micro distribution. Real domain alignment is comprehensive alignment. In order to achieve all-round alignment.

In this paper, this study rethink "what is real domain alignment". This study think that the real alignment is the alignment of all aspects of the domain, including the meso and micro level. Based on this thinking, this study proposes a high confidence pseudo label adaptive method of double classification (DCP). This study uses double classifiers to complete this task and achieve global distribution alignment Generative Adversarial Networks (GAN); clustering network is used as auxiliary network regularization meso and micro distribution. In addition, in the process of false labeling, false pseudo-labeling will cause a chain reaction, so This study selectively labels the samples. The contributions of this work can be summarized as follows:
- Rethink the adaptation between different domains. This study think that alignment should not only be global alignment, but also be meso and micro alignment.
- This study proposes a double classifier method based on high confidence labels. Methods by aligning the centroid distance and centroid sample distance of the two classifiers, the comprehensive alignment of different domains was realized.
- High confidence pseudo-labeling method is proposed to reduce the false pseudo-labeling error of dual network.
- Through comparative experiments, the results demonstrate the proposed approach can achieve new state-of-the-art performance on benchmark datasets

## II. Related work

In this section, this study will introduce the related work in three aspects: unsupervised domain adaptation, pseudo label, double classifiers.

### A. Unsupervised domain adaptation

Many unsupervised Da methods have been developed and applied to cross domain applications, such as object recognition, object detection, semantic segmentation and so on. At present, the most popular methods are feature-based, sample-based, and mode-based. Feature-based methods mainly include several branches, including measurement based on distribution discrepancy, clustering, alignment, matching, selection, etc. For example: To feature-based method, [27] reduce the distance between different domains in the feature space. To based-samples, [28] give different weights to different samples. Model based learning mainly includes based deep learning and based generative adversarial networks. Based deep learning always utilized with feature-based together now. [15, 29] reduce the difference of feature distribution through deep learning feature extraction. for example, based on features. For example, based on the generative adversarial networks, [3] uses discriminators to narrow the differences of domain attributes. These methods all contribute to the domain adaptation, but they mainly focus on the alignment of global distribution, which is often not enough for visual tasks.

### B. pseudo label

In the process of local alignment, samples need to be labeled, which is called pseudo-label. There is a common problem in the process of pseudo-labeling, that is, wrong pseudo-labels[30]. Ideally, this study wants the labeled sample label to be the real label of the sample, so that it won't cause a chain error. Current experiments show that false tags are widespread. In order to alleviate this problem, [14] use the classifier trained in the source domain to label directly; [31] use the clustering method to classify the samples with similar features; [32] use the centroid method to label according to the distance from the sample to the centroid; [22] improve the threshold requirements of labeled samples, so as to reduce the labeling errors; [33] use condition generative adversarial networks to eliminate domain migration domain shift noise, improve the robustness of the tag. These methods reduce pseudo-label errors to a certain extent. However, the correctness of the pseudo-label has a great influence on the regularization of the subsequent sample distribution. So, this study needs to take a more rigorous labeling strategy. In this paper, the double classifiers high confidence label method is used to improve the accuracy of labeling.

### C. Subdomain Adaptation

In the field of unsupervised domain adaptation, recent years lots of unsupervised domain adaptation methods have been developed and successfully used in cross-domain application. According to the existing results, it is not difficult to find that there are many differences between the early domain adaptation method and the recent domain adaptation method. The focus of the early domain adaptation method is to align the global distribution. These methods mainly focus on the macro characteristics of data sets. With the deepening of research and the thinking of vision related problems, it is found that macro alignment cannot help the network to extract specific class features. It is difficult to improve the performance of classifier. Therefore, while reducing the global difference of the domain, this study should extract the features of the region classification as much as possible. Some of the latest works focus on few local differences, such as using clustering network as an auxiliary model to explore the clustering information of samples. For example, [14] reduce the MMD difference of small domains, for example, [22] reduce the false mark error by optimizing the false mark strategy. Then, these methods reduce the differences between subdomains to a certain extent. However, these methods only focus on the differences between sub domains, and ignore the structural features between classes and samples. And these characteristics have a great relationship with the distribution of sample characteristics. Therefore, in the process of aligning the global distribution, this study needs to combine the meso and micro distribution characteristics of samples.

## III. METHOD

In the following subsections, this study gives the formulation of the problem, describe each component of the proposed method in detail, including generative adversarial network, clustering network aided distribution regularization and High-confidence pseudo label.

### A. Preliminaries

**Problem Formulation**: In unsupervised domain adaptation, This study are given a labeled source domain $D_s = \{x_i^s, y_i^s\}_{i=1}^{N_s} D_s = \{x_i^s, y_i^s\}_{i=1}^{N_s}$ and a unlabeled target domain $D_t = \{x_i^t\}_{i=1}^{N_t}$ where $N_s$ and $N_t$ denote the size of $D_s$ and $D_t$, respectively. $D_s$ and $D_t$ are sampled from different data distributions $P_s(\mathcal{X}_s|\mathcal{Y}_s)$, $P_t(\mathcal{X}_t|\mathcal{Y}_t)$, respectively. Our goal is to design a neural network $\mathbf{y} = f(\mathbf{x})$ that formally reduces the shifts in the distributions of the relevant domains and ensure the knowledge learned from the source domain is well generalized in the target domains. Then, this study introduces DCP algorithm.

**Generative Adversarial Networks** Adversary generated network is a popular network structure recently, which has a good effect in aligning the distribution of different domains. Adversary generated network mainly includes generator (G), domain discriminator (D) and classifier (C). The generator mainly uses the initial data to extract features, and constantly optimizes the generated features to make the discriminator unable to identify the source of the domain; the purpose of domain identification is to identify the domain to which the sample belongs, so as to optimize the discriminating ability of the discriminator. The loss of the generated countermeasure network consists of three parts.

$$\mathcal{L}_{adv} = \mathcal{L}_G + \mathcal{L}_D + \mathcal{L}_{C1} \qquad (1)$$

Where $\mathcal{L}_G$ is the generator loss, $\mathcal{L}_D$ is the discriminator loss, $\mathcal{L}_{C1}$ is the classification loss of the source domain。 $\mathcal{L}_D$ is calculated as follow:

$$\mathcal{L}_D = \mathbb{E}_{\mathbf{x}^s \sim P_s} \log[D(g_s)] + \mathbb{E}_{\mathbf{x}^t \sim P_t} \log[1 - D(g_t)] \qquad (2)$$

$\mathcal{L}_{C1}$ is the cross entropy classification loss of the source domain, and the calculation method is as following?

$$\mathcal{L}_{c1} = \mathbb{E}_{(\mathbf{x}^s,\mathbf{y}^s) \sim P_s} \mathcal{L}(C(f^s), y^s) \qquad (3)$$

As the optimization objective of discriminator is opposite to that of generator, the optimization process is iterative optimization, and the specific process is given later.

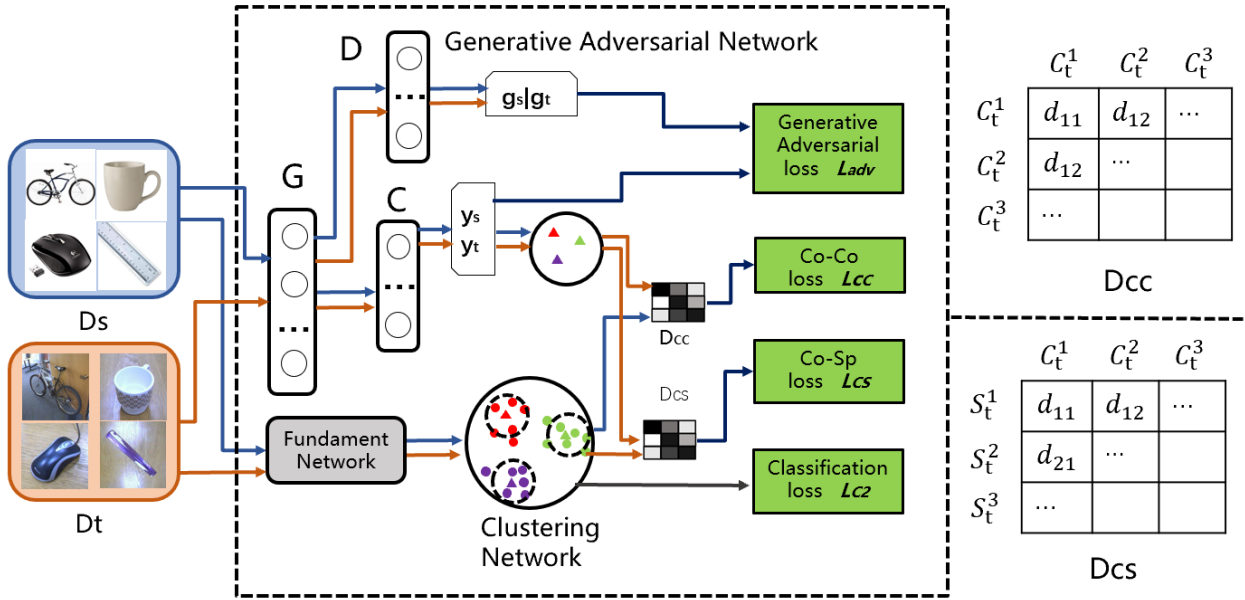

Fig. 2. The proposed framework of DCP network. The framework consists of a generative adversarial network and a clustering network. Two network base pseudo labels utilize centroids and samples generate $D_{cc}$ and $D_{cc}$. Through the distance between the matrices, the meso distribution and micro distribution are regularized.

## B. Clustering network aided distribution regularization

In general, clustering algorithms help to optimize the feature distribution of samples by combining pseudo label and centroid to extract the clustering features of the target domain. But it is easy to ignore the relationship between centroids and the relationship between centroids and samples in the process of optimization. Such consideration is of practical significance, for example, the centroids of stools and chairs are distributed closer in the feature space than those of stools and cups; for chairs, images with slightly different angles of the same chair are distributed closer in the feature space than images with completely different shapes and angles. So, based on this idea, this study uses double networks to align the meso and micro distribution of samples.

The clustering network is another kind of dual network in this paper. The feature extraction of F is completed through the basic network $f(x_i^s)$, Then, the K-means clustering algorithm is used to output the classification $y_{cl}$ of samples, and the cross entropy of source samples $\mathcal{L}_{C2}$ is used to classify samples to complete the network training optimization.

Using the classification results, this study calculates the centroids of different classes. For the k-class source domain, the initial centroids are calculated as follows:

$$\overline{co}_K^{s,0} = \frac{\sum_{i=1}^{n_s} f(x_i^s)(y_{cl} \in K)}{\sum_{i=1}^{n_s} 1(y_{cl} \in K)} \quad (4)$$

Based on the centroid of the class, the centroid distance D is obtained by using the centroid distribution $d_{cc}$. It should be noted that the distance here is relative distance, not absolute distance. Then, the centroid distance matrix $D_{cc}$ is constructed based on the centroid distance $d_{cc}$, in which each element $d_{ij}$ represents the distance between the centroid of class i and the centroid of class j. Using the same calculation method, this study calculates the centroid $co'^{s,0}_K$ and the centroid distance $d'_{cc}$ based on the classification result $y_s$ of GAN and construct the centroid distance matrix $D'_{cc}$. Finally, this study proposes the centroid distance loss $\mathcal{L}_{CC}$, using the distance between the centroids to lose the distance between the rule centroids, so as to align the mesoscopic distribution of different fields. The calculation method is as follows:

$$\mathcal{L}_{CC} = \frac{1}{K^2} \sqrt{\sum_{i=1}^{K} \sum_{j=1}^{K} (D'_{CC}(i,j) - D_{CC}(i,j))^2} \quad (5)$$

Similarly, for the micro perspective, that is, for different samples of the same class, this study starts from the distance between the centroid and the samples. The distance $d_{cs}$ between centroid and sample eigenspace is calculated, and the centroid sample matrix $D_{cs}$ is also constructed based on $d_{cs}$. the centroid sample matrix $D'_{cs}$ of GAN is constructed by the same method. Finally, the centroid sample distance loss L is proposed $\mathcal{L}_{CS}$. Based on small batch samples, the micro distribution of samples is regularized by using the distance between samples and centroid.

$$\mathcal{L}_{CS} = \frac{1}{K*N_b} \sqrt{\sum_{i=1}^{K} \sum_{j=1}^{N_b} (D'_{CS}(i,j) - D_{CS}(i,j))^2} \quad (6)$$

Where K is the number of classes and $N_b$ is the number of samples.

## C. High-confidence pseudo-label

In the process of clustering regularization, our method needs to label the samples. There are two ways to label, one is to label samples by clustering network, the other is to label samples by confrontation network classifier. Recent studies[34] have shown that the accuracy of pseudo-labels in the early stage is not high. The wrong pseudo-label will lead to centroid deviation and gradient optimization error, that is to say, the wrong false mark will lead to chain error. Therefore, the result of false pseudo-label is worse than that of no pseudo-label. Based on this thinking, this study adopts a high standard selective pseudo labeling strategy.

Combined with the actual situation of pseudo-labeling, generally, the pseudo-labeling error rate is higher in the early stage, so more strict requirements are needed in the early stage. Then, in the process of loop optimization, the pseudo-labeling accuracy increases, the pseudo-labeling requirements are reduced, and the new samples are labeled. This paper proposes a high confidence pseudo-labeling method, which is adaptive threshold labeling. The classification results of generative adversarial network generation network and clustering network are used to screen and label small batch samples. The requirement of screening is to reach a certain threshold. The requirement of screening is to reach a certain threshold. The threshold proposed in this paper is not a fixed value, but a variable threshold percentage. Two classifiers adopt different threshold percentage, and the threshold percentage of generative adversarial network is $\tau_{adv}$, and the threshold percentage of clustering network is $\tau_{clu}$. In this way, through double threshold pseudo-labeling, the requirement of pseudo-labeling is greatly improved. In the same batch of samples, only the samples which are close to the centroid in the double classification results are labeled, so the pseudo-labeling accuracy will be greatly improved. Among them, the percentage of pseudo-label threshold of generative adversarial network is $\tau_{adv}$ is calculated as follows:

$$\tau_{adv} = \frac{1}{(1+e^{(-0.0001*T*T)})} - 0.1 \quad (7)$$

Where $T$ is the number of iterations and $\tau_{adv}$ is the curve about the rise of $T$. For clustering network, the curve trend of $\tau_{adv}$ is similar to that of $\tau_{clu}$, but considering that the accuracy of clustering network is relatively high in the early stage. This study selects a higher proportion in the early stage, and the calculation method is as follows:

$$\tau_{clu} = \frac{1}{(1+e^{(-0.01*T)})} \quad (8)$$

It should be noted that: in order to prevent the network from selecting easy classes for labeling, each class is allocated equally in proportion to the number of tags.

## IV. EXPERIMENTS

In this paper, this study evaluates our method on object recognition and digit classification. The four data sets, including Digits, office, office-31, Visda-2017. Firstly, our approach is compared with state-of-the-art UDA methods to evaluate its effectiveness. An ablation study is conducted to demonstrate the effect of different components.

## A. Datasets description

**Digits** is common data set that focuses on digit recognition. This study follow the protocol of [35] and utilize three subsets: SVHN(S), MNIST(M), and USPS(U). MNIST contains gray digits images of size 28*28, USPS contains 16*16 gray digit, and SVHN contains color 32*32 digits images that might contain more than one digit in each image. This study build three transfer tasks: MNIST to USPS (M→U), USPS to MNIST (U→M) and SVHN to MNIST (S→M).

**Office-31**[36] is a popular benchmark dataset for visual domain adaptation tasks, which contains images of 31 object categories originating from office environment in three distinct datasets: Amazon (A), Webcam (W) and DSLR (D). There are a total of 4110 images with an average of 90 images per category for Amazon, 26 images per category for Webcam, and 16 images per category for DSLR. This study carries out the experiments on six source–target domain pairs, i.e., A →D, A → W, D → A, D → W, W → A, and W → D.

**Office-home**[37] is a more complicated dataset than Office-31, which contains around 15,500 images from 65 categories. Four extreme distinct domains exist in the Office-Home dataset: Artistic images (A), Clip Art (C), Product images (P), and Real-World images (R). In this paper, this study evaluates DCP on all twelve transfer tasks.

**ImageCLEF-DA** is a benchmark data set for ImageCLEF-DA 2014 domain adaptation challenge. ImageCLEF-2014 dataset consists of three datasets: Caltech-256 (C), ILSVRC 2012 (I), and Pascal VOC 2012 (P). There are 12 common classes and total 600 images in each domain. This study use all domain combinations and build six transfer tasks: I → P, P → I, I → C, C → I, C → P, P → C.

**VisDA-2017**[38] is a challenging simulation-to-real dataset that mainly focuses on the 12-class synthesis-to-real object recognition task. The source domain contains 152 thousand synthetic images generated by rendering 3D models while the target domain has 55 thousand real object images sampled It contains over 280k images across 12 classes in the training, validation, and test domains.

## B. Baseline methods

For vanilla unsupervised DA in digit recognition, This study compare DCP with ADDA[3], ADR[5], CDAN[7], MCD[9]. For object recognition, This study compare DCP with ResNet50[2], DANN[1], DAN[4], ADDA[3], CDAN[7], JAN[15], GTA[24]. Note that results are directed cited from published papers.

## C. Implementation details

For domain adaptation tasks on the Office-31, Office-Home and VisDA-2017 datasets, This study adopt the pre-trained ResNet-50[2] and ResNet-101[2] as the fundament network; here, the momentum and learning rate of mini-batch SGD optimizers are set to 0.9 and 0.003. For domain adaptation tasks on digit datasets, this study uses the LeNet-5 as the fundament network; here, the mini-batch SGD optimizer is set to have a batch size of 36, the momentum of 0.5 and learning rate of 0.01. For all these datasets, the regularization parameter α is set to 0.1.

## D. results

### 1) Digit Classification

The classification results of three tasks of MNIST–USPS–SVHN are shown in Table 1. DCP largely outperforms all baselines, In particular, S → M has achieved the best recognition effect, the accuracy is 97%; from the average point of view, the accuracy of DCP has been improved by 1.2% compared with other latest methods.

Overall, DCP achieves better average accuracy and more stable results with lower standard error.

### 2) Object Recognition

The classification results of Office-Home, Office-31, ImageCLEF-DA, and VisDA-2017 are, respectively, shown in Tables 2–5. DCP outperforms all compared methods on most transfer tasks. In particular, DSAN substantially improves the average accuracy by large margins (more than 2%) on Office-Home, ImageCLEF-DA and VisDA-2017. As shown in Table 2, in the medium-sized dataset office home, our method has been improved in almost every task, and basically achieved the best results in 12 subtasks. This may be because our extracted features retain more meso and micro distribution of samples. In the office-31 dataset, our method can achieve the same effect as the latest method. In the image clef-da dataset, our method achieves the best results on four subtasks. In C → P, compared with the latest method, the accuracy is improved by 4%. In the large data set visda-2017, This study continues to maintain the same effect as the latest methods on some simple classes and achieve the best effect on some more difficult to identify classes, and far more than other methods, such as skateboard and truck. The encouraging results indicate the importance of subdomain adaptation and show that DSAN is able to learn more transferable representations.

TABLE I
ACCURACY (%) ON DIGIT DATASET FOR UNSUPERVISED DOMAIN ADAPTATION

| Method | S→M | U→M | M→U | Avg |
|---|---|---|---|---|
| DANN[1] | 71.1±0.1 | 73.0±2.0 | 77.1±1.8 | 73.7 |
| ADDA[3] | 76.0±1.8 | 90.1±0.8 | 89.4±0.2 | 73.0 |
| ADR[5] | 95.0±1.9 | 93.1±1.3 | 93.2±2.5 | 93.8 |
| CDAN[7] | 89.2 | **98.0** | 95.6 | 94.3 |
| MCD[9] | 96.2 | 94.1 | 94.2 | 94.8 |
| DSAN[14] | 90.1±0.4 | 95.3±0.1 | **96.9**±0.2 | 94.1 |
| DCP(ours) | **97.1**±0.2 | 94.8±0.2 | 96.3±0.1 | **96.0** |

The experimental results further reveal several insightful observations.

improvement from previous global domain adaptation methods to subdomain adaptation methods is crucial for

TABLE II
ACCURACY (%) ON OFFICE-HOME DATASET FOR UNSUPERVISED DOMAIN ADAPTATION

| Method | A→C | A→P | A→R | C→A | C→P | C→R | P→A | P→C | P→R | R→A | R→C | R→P | Avg |
|---|---|---|---|---|---|---|---|---|---|---|---|---|---|
| ResNes50[2] | 34.9 | 50.0 | 58.0 | 37.4 | 41.9 | 46.2 | 38.5 | 31.2 | 60.4 | 53.9 | 41.2 | 59.9 | 46.1 |
| DAN[4] | 43.6 | 57.0 | 67.9 | 45.8 | 56.5 | 60.4 | 44.0 | 43.6 | 67.7 | 63.1 | 51.5 | 74.3 | 56.3 |
| DANN[1] | 45.6 | 59.3 | 70.1 | 47.0 | 58.5 | 60.9 | 46.1 | 43.7 | 68.5 | 63.2 | 51.8 | 76.8 | 57.6 |
| CDAN[7] | 50.6 | 65.9 | 73.4 | 55.7 | 62.7 | 64.2 | 51.8 | 49.1 | 74.5 | 68.2 | 56.9 | 80.7 | 62.8 |
| CDAN+E[7] | 50.7 | 70.6 | 76 | 57.6 | 70.0 | 70.0 | 57.4 | 50.9 | 77.3 | 70.9 | 56.7 | **81.9** | 65.8 |
| JAN[15] | 45.9 | 61.2 | 68.9 | 50.4 | 59.7 | 61.0 | 45.8 | 43.4 | 70.3 | 63.9 | 52.4 | 76.8 | 58.3 |
| SAFN[13, 17] | 52.0 | 71.7 | 76.3 | 64.2 | 69.9 | 71.9 | 63.7 | 51.4 | 77.1 | 70.9 | 57.1 | 81.5 | 67.3 |
| DCP(ours) | **56.2** | **73.2** | **79.8** | **66.5** | **77.4** | **73.6** | **64.5** | **53.8** | **79.6** | **71.9** | **58.1** | **81.9** | **69.7** |

TABLE III
ACCURACY (%) ON OFFICE-31 DATASET FOR UNSUPERVISED DOMAIN ADAPTATION

| Method | A→D | A→W | D→A | D→W | W→A | W→D | Avg |
|---|---|---|---|---|---|---|---|
| ResNes50[2] | 68.9±0.2 | 68.4±0.2 | 62.5±0.3 | 96.7±0.1 | 60.7±0.3 | 99.3±0.1 | 76.1 |
| DAN[4] | 78.6±0.2 | 80.5±0.4 | 63.6±0.3 | 97.1±0.2 | 62.8±0.2 | 99.6±0.1 | 80.4 |
| DANN[1] | 79.7±0.4 | 80.2±0.4 | 68.2±0.4 | 96.9±0.2 | 67.4±0.5 | 99.1±0.1 | 82.2 |
| ADDA[3] | 77.8±0.3 | 86.2±0.5 | 69.5±0.4 | 96.2±0.3 | 68.9±0.5 | 98.4±0.3 | 82.9 |
| CDAN[7] | 89.8±0.3 | 93.1±0.2 | 70.1±0.4 | 98.2±0.2 | 68.0±0.4 | **100.0±0.0** | 86.6 |
| JAN[15] | 84.7±0.3 | 85.4±0.3 | 68.6±0.3 | 97.4±0.2 | 70.0±0.4 | 99.8±0.2 | 84.3 |
| GTA[24] | 90.8±1.8 | 94.4±0.1 | 72.2±0.6 | 98.0±0.2 | 70.2±0.1 | **100.0±0.0** | 87.6 |
| BSP[26]+CDAN | **93.0±0.2** | 93.3±0.2 | **73.6±0.3** | 98.2±0.2 | 72.6±0.3 | **100.0±0.0** | **88.5** |
| **DCP(ours)** | 91.6±0.2 | **95.3±0.1** | 73.1±0.2 | **98.3±0.4** | **72.7±0.3** | **100.0±0.0** | **88.5** |

TABLE IV
ACCURACY (%) ON IMAGECLEF-DA DATASET FOR UNSUPERVISED DOMAIN ADAPTATION

| Method | C→I | C→P | I→C | I→P | P→C | P→I | Avg |
|---|---|---|---|---|---|---|---|
| ResNes50[2] | 78.0±0.2 | 65.5±0.3 | 91.5±0.3 | 74.8±0.3 | 91.2±0.3 | 83.9±0.1 | 80.7 |
| DAN[4] | 84.1±0.4 | 69.8±0.4 | 93.3±0.2 | 75.0±0.4 | 91.3±0.4 | 86.2±0.2 | 83.3 |
| DANN[1] | 87.0±0.5 | 74.3±0.5 | 96.2±0.4 | 75.0±0.6 | 91.5±0.5 | 86.0±0.3 | 85.0 |
| CDAN[7] | 90.5±0.4 | 74.5±0.3 | 97.0±0.4 | 76.7±0.3 | 93.5±0.4 | 90.6±0.3 | 87.1 |
| JAN[15] | 89.5±0.3 | 74.2±0.3 | 94.7±0.2 | 76.8±0.4 | 91.7±0.3 | 88.0±0.2 | 85.8 |
| CDAN[7] | 90.5±0.4 | 74.5±0.3 | **97.0±0.4** | 76.7±0.3 | 93.5±0.4 | 90.6±0.3 | 87.1 |
| DCP(ours) | **92.4±0.2** | **78.6±0.1** | 95.7±0.3 | **78.8±0.2** | **95.7±0.3** | **92.8±0.2** | **89.0** |

TABLE V
ACCURACY (%) ON VISDA-2017 DATASET FOR UNSUPERVISED DOMAIN ADAPTATION

| Method | air | bic | bus | car | horse | knife | mot | pers | plant | skt | train | truck | Avg |
|---|---|---|---|---|---|---|---|---|---|---|---|---|---|
| ResNes101[2] | 72.3 | 6.1 | 63.4 | 91.7 | 52.7 | 7.9 | 80.1 | 5.6 | 90.1 | 18.5 | 78.1 | 25.9 | 49.4 |
| DAN[4] | 68.1 | 15.4 | 76.5 | 87.0 | 71.1 | 48.9 | 82.3 | 51.5 | 88.7 | 33.2 | 88.9 | 42.2 | 62.8 |
| DANN[1] | 81.9 | **77.7** | 82.8 | 44.3 | 81.2 | 29.5 | 65.1 | 28.6 | 51.9 | 54.6 | 82.8 | 7.8 | 57.4 |
| JAN[15] | 75.7 | 18.7 | 82.3 | 86.3 | 70.2 | 56.9 | 80.5 | 53.8 | 92.5 | 32.2 | 84.5 | 54.5 | 65.7 |
| DSAN[14] | 90.9 | 66.9 | 75.7 | 62.4 | 88.9 | 77.0 | **93.7** | **75.1** | **92.8** | 67.6 | **89.1** | 39.4 | 75.1 |
| MCD[9] | 90.3 | 62.6 | **84.8** | 71.7 | 85.9 | 72.9 | **93.7** | 71.9 | 86.8 | **79.1** | 81.6 | 14.3 | 74.6 |
| DCP(ours) | **91.3** | 70.5 | 77.3 | **75.4** | **89.7** | **78.3** | 90.5 | 71.2 | 91.5 | 70.2 | 88.8 | **57.2** | **97.3** |

1) In standard domain adaptation, subdomain adaptation methods (DAN[4], CDAN[7], and our DCP) outperform previous global domain adaptation methods. The domain adaptation; previous methods align global distribution without considering the relationship between subdomains, whereas DSAN accurately aligns the relevant subdomain distributions, which can capture more fine-grained information for each category.

2) Our method is based on counterwork network and uses double classifiers. Compared with Generative Adversarial Networks method[15], our method achieves better results in most sub tasks.

3) In addition, this study notice that our method is more effective on medium and large datasets (such as Office-Home and VisDA-2017) than on small datasets. Based on this, this study speculate that a larger data set can better reflect the meso and micro distribution of data, so our method also achieves good results for the tasks that are difficult to distinguish by conventional methods.

## V. CONCLUSION

In this paper, aiming at unsupervised domain adaptation, this study rethink what is real domain adaptation. Different from previous domain adaptations, which focus on global distribution and micro distribution, this study think that real domain adaptations should be aligned in global, meso and micro space. Based on these considerations, this study proposes a high confidence labeling method based on double classifiers, which combines centroid and sample distribution to align meso and micro distribution. In addition, this study also proposes a high confidence pseudo marking method, which can effectively reduce the error marking. Extensive experiments conducted on both object recognition and digit classification benchmark datasets demonstrate the effectiveness of our method.

Our work has achieved good results in digital recognition and target recognition tasks, but our method has some limitations in speed and few-shot learning. Due to the need to align the meso and micro distribution of samples, the calculation is large, so the speed is slow. In addition, because our method aligns the features of different levels in different domains, it needs more samples to have good feature extraction effect.

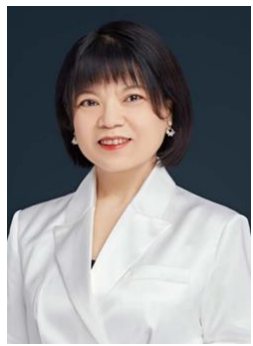

**Li Li** received the B.S., M.S. degrees in electrical automation from Shenyang Agriculture University, Shenyang, China, in 1986, 1993, and 2000, respectively, and the Ph.D. degree in mechatronics engineering from Shenyang Institute of Automation, Chinese Academy of Science, in 2003. She joined Tongji University, Shanghai, China, in 2003, and is professor of Control Science and Engineering.

Her research interests are in production planning and scheduling, computational intelligence, semiconductor manufacturing, and energy systems.

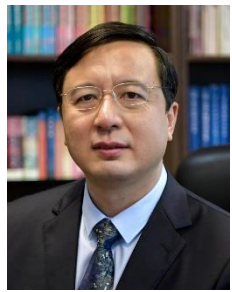

**Jie Chen** (Fellow, IEEE) received the B.S., M.S., and Ph.D. degrees in control theory and control engineering from the Beijing Institute of Technology, Beijing, China, in 1986, 1993, and 2000, respectively.

He is currently a Professor of Control Science and Engineering with the Beijing Institute of Technology. His current research interests include intelligent control and decision in complex systems, multiagent systems, and optimization methods.

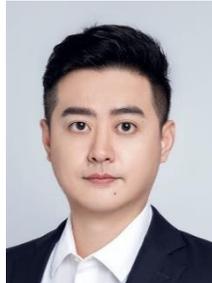

**Kuo-Yi Lin** is currently an Associate Professor of Department of Control Science and Engineering, Tongji University. He received the B.S. degree in Statistics from Cheng Kung University, Taiwan, China, in 2007 and the M.S. degree and Ph.D. degree in Industrial Engineering and Engineering Management from Tsing Hua University, Taiwan, China, in 2009 and 2014. He is director of China Excellent Business Decision Making Society, member of Intelligent Simulation Optimization and Scheduling Committee Of China Simulation Society, member of Natural Computing And Digital Intelligent City Committee Of China Artificial Intelligence Society, member of Industrial Big Data And Intelligent System Branch Of China Mechanical Engineering Society. He is mainly engaged in Intelligent Manufacturing, Federated Learning, Quantum Algorithm and Transfer Learning.

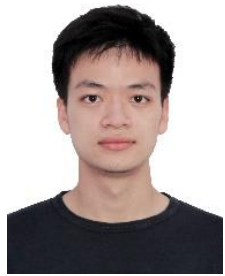

**Huihuang Chen** received the bachelor's and master's degrees from Harbin Engineering University, Harbin, China, in 2016 and 2018, respectively.

His current research interest includes transfer learning.